\def\year{2021}\relax
\documentclass[letterpaper]{article} 
\usepackage{aaai21}  
\usepackage{times}  
\usepackage{helvet} 
\usepackage{courier}  
\usepackage[hyphens]{url}  
\usepackage{graphicx} 
\usepackage{natbib}  
\usepackage{caption} 
\title{Human Capabilities as Guiding Lights for the Field of AI-HRI:\\ Insights from Engineering Education}
\author {
    Tom Williams and Ruchen Wen\\
}

\affiliations{
    MIRRORLab\\Colorado School of Mines\\
    Golden, CO 80401\\
    \{twilliams,rwen\}@mines.edu
}

\begin{document}

\maketitle

\begin{abstract}
Social Justice oriented Engineering Education frameworks have been developed to help guide engineering students' decisions about which projects will genuinely address human needs to create a better and more equitable society. 
In this paper, we explore the role such theories might play in the field of AI-HRI, 
consider the extent to which our community is (or is not) aligned with these recommendations, and envision a future in which our research community takes guidance from these
theories. In particular, we analyze recent AI-HRI (through analysis of 2020 AI-HRI papers) and consider possible  futures of AI-HRI (through a speculative ethics exercise). Both activities are guided through the lens of the Engineering for Social Justice (E4SJ) framework, which centers contextual listening and enhancement of human capabilities. Our analysis suggests that current AI-HRI research is not well aligned with the  guiding principles of Engineering for Social Justice, and as such, does not obviously meet the needs of the communities we could be helping most. As such, we suggest that motivating future work through the E4SJ framework could help to ensure that we as researchers are developing technologies that will actually lead to a more equitable world.
\end{abstract}

\section{Introduction}

The AAAI Code of Ethics begins with two key principles to help guide the practice (in and beyond the research laboratory) of AI Professionals: ``An AI professional should... (1) Contribute to society and to human well-being, acknowledging that all people are stakeholders in computing''; and ``(2) Avoid harm''~\cite{aaai2019aaai}. While these principles are admirable goals, there are well known challenges to effective use of such ethical codes~\cite{giorgini2015researcher}. 
For example, the apparent difficulty of ensuring that the research published at our professional conferences truly benefits society and truly causes minimal harm is evidenced by the wide variety of new surveillance technologies published at AI conferences each year despite widespread discussion of their harms in our community and in the popular press, especially to minoritized and oppressed communities~\cite{gebru2020race}. Moreover, even the goal of avoiding harm may be called into question, as humans deserve more than simply to avoid being harmed. As such, codes of ethics instruct engineers to avoid harm but do not tell engineers what to do to build a better world~\cite{harris2008good}.
Instead of relying on ineffective professional society codes to safeguard our communities, we instead argue that if we desire our communities to produce research that genuinely contributes to society and genuinely avoids harm, we should work to cultivate a research ecosystem that prioritizes socially beneficial outcomes, and that properly recognizes computing as what it has become: a social science~\cite{connolly2020computing}.

For better or worse, this is not a unique challenge. Engineering educators at both the K-12~\cite{mcgowan2020engineering} and collegiate levels~\cite{swartz2019sociotechnical} struggle each year to cultivate in their students an understanding of the sociotechnical dimensions of engineering, so that students understand engineering as not simply a process of applying mathematical tools to derive solutions to abstract problems handed down from on high, but rather as a process of developing and deploying technologies that attending to the real needs of real people in order to genuinely better our society. 

This is of particular relevance to the AI-HRI community due to the fine line between AI-HRI science and AI-HRI engineering~\cite{depalma2014quis}, and the 
close relationship between robotics more generally and engineering. As a whole, the field of robotics is one organized around specific engineering efforts. And even within the subfield of HRI, which includes a large number of researchers from the social and behavioral sciences, and which has historically been dominated by experimental research, experiments are typically motivated in terms of the design recommendations they facilitate, i.e., the ways that they can advance robotics engineering practice. 

As such, we argue that to understand how best to steer our community toward the development of AI-HRI solutions that are truly of benefit to society, we should be looking not only to fields such as design studies, value-sensitive design~\cite{friedman1996value}, and other areas already popular in the HCI community, but also to the Engineering Education community. 

Just as AI Ethicists have been working to move beyond theories of AI Ethics grounded in Moral Philosophy or Fairness, Accountability, and Transparency in favor of theories that center notions of power and justice~\cite{bennett2020point,le2020we}, researchers from the Engineering Education literature have been exploring critical pedagogies that place Social Justice as the central goal of engineering education~\cite{leydens2014design,leydens2017engineering,nieusma2013engineering,riley2008engineering,winberg2017using}.
In this work, we specifically consider the \textit{Engineering for Social Justice} (E4SJ) framework delineated by \citet{leydens2017engineering} in their pioneering 2017 book ``Engineering Justice'' (see also~\citet{leydens2017confronting,leydens2016making}), and show how by critiquing recent AI-HRI work through this framework we can reveal productive paths forward towards creating intelligent, interactive robotic technologies that are likely to be of true societal benefit.

\subsection{Engineering for Social Justice}
Engineering for Social Justice (E4SJ) is a set of engineering practices which strive to enhance
human capabilities through equitable distribution of opportunities and resources while reducing imposed risks and harms within specific communities~\cite{leydens2017engineering}.
Specifically, when viewed through the lens of the E4SJ framework, the central goal of engineering is that engineers should work alongside communities to develop engineering solutions that \textbf{enhance human capabilities} in a way that aligns with community priorities. The way that this goal should be pursued is then specified according to five supplemental criteria:

\begin{enumerate}
    \item \textbf{Listening contextually} -- at the basis of Engineering practice is listening to and empathizing with different communities' perspectives and their constituent struggles, concerns, desires, and preferences.
    \item \textbf{Identifying structural conditions} -- these perspectives must be understood through the lens of the structural conditions (e.g., racial, gendered, socioeconomic) that constrain those communities' opportunities, desires, and aspirations, as well as the structural conditions that constrain the engineers own opportunities, desires, and aspirations.
    \item \textbf{Acknowledging political agency / mobilizing power} -- engineers must understand how communities' political power and agency (as well as their own) can be mobilized and leveraged when developing engineering solutions.
    \item \textbf{Increasing opportunities and resources} -- engineers should work with communities to identify the opportunities (e.g. health, education, housing, and employment) could be improved by leveraging and mobilizing political power, engineering solutions, and other resources, as mediated by structural conditions.
    \item \textbf{Reducing imposed risks and harms} -- engineers should work with communities to identify how to leverage the identified resources to develop solutions in a way that is sensitive to how the solution's potential risks that will be distributed across the community. 
\end{enumerate}

This framework, while originally designed as a framework for cultivating more equitable practice in the broad context of engineering education, may be applied to the fields of AI and Robotics as a form of third-wave AI Ethics~\cite{bennett2020point,le2020we} (i.e., focused not solely on moral philosophy or fairness, accountability, and transparency, but rather moreso on issues of power and social justice). Taking this approach requires using a community-focused approach, in which the designer or researcher specifies and focuses on a particular community that is structurally disadvantaged in particular ways, such as children, AAC device users, farmers, LGBT+ people, Black people, women, immigrants, incarcerated people, and so forth. Taking this community-focused approach is critical to the development of technology from this perspective. Technologies need to be designed with specific communities in mind because different communities have inherently opposed values and goals. Navigating this tension requires developing an awareness of and clearly specifying the community you are designing for, and then developing a local understanding of that community (or, at least, explicitly building on the findings of others who have developed and documented this level of understanding). Failure to do so risks technology only meeting the values and needs of the technology's developer. Similarly, developing this local understanding forces the technology developer to ensure that the technology fits into the lives of the actual people who are expected to use it.

When viewed through this lens, two key elements stand out to us as immediately applicable to research practice in AI-HRI: (1) the research we perform should be motivated by the needs of \textbf{specific communities} in order to help those communities achieve an equitable distribution of opportunities and resources that they otherwise lack due to structural conditions (cf.~\citet{leydens2014design}), and (2) the types of solutions proposed in our research should seek to achieve this goal specifically through means that advance key human capabilities. In particular, Leydens and Lucena suggest designing to advance one or more of the 10 capabilities delineated by Martha Nussbaum, which we discuss in the following section.

\subsection{Human Capabilities}

Nussbaum introduced a human development paradigm, which applies the capabilities approach, and generated a list of central human capabilities from the following ten aspects, which we re-describe so as to be intuitive to our field~\cite{nussbaum2009creating}:

\begin{enumerate}
    \item \textbf{Life} -- Ability to live to the end of a life of normal human length without it being cut short.
    \item \textbf{Bodily health} -- Ability to live in good health, with access to adequate nutrition and shelter. 
    \item \textbf{Bodily integrity} -- Ability to move freely, be free from assault, and have sexual satisfaction and reproductive choice.
    \item \textbf{Senses, imagination, and thought} -- Ability to use senses to imagine, think, and reason, 
    informed by adequate education,
    to produce and experience works, events, political and artistic speech, and religious exercise, of one's own choice; and the ability to have pleasurable experiences and avoid pain
    \item \textbf{Emotions} -- Ability to experience and explore positive and justified negative emotions and feelings towards others. 
    \item \textbf{Practical reason} -- Ability to develop and engage in reflection as to what is ``good'', and use the resulting personal axiologies to engage in goal-driven self-reflection. 
    \item \textbf{Affiliation} -- Ability both to (a) live, engage socially, and empathize with others; and (b) be treated with dignity and respect, and avoid discrimination.
    \item \textbf{Other Species} -- Ability to live with and experience concern for other species (e.g. plants and animals) and the natural world in general.
    \item \textbf{Play} -- Ability to laugh and play.
    \item \textbf{Control over one's political and material environment} -- 
    Ability to participate effectively in political processes affecting one's life, hold property, seek employment and work in a human-like, goal-driven, and social way.
\end{enumerate}

There are important questions that can be raised about the ontology (what ``is'') of capabilities, such as who gets to decide what capabilities belong in such a taxonomy. 
There are important questions that can be raised about the epistemology (what ``is known'') of capabilities, such as how we adjudicate ``levels'' of capabilities, when those levels have been reached, and who gets to decide on those levels.
And there are important questions that can be raised about the axiology (what ``is valued'') of capabilities, such as how different capabilities might be differentially valued across different communities and in different cultures. Nevertheless, these capabilities may yet serve as a starting point for a capability-directed discussion of our own field's advancement of human capabilities. And while the E4SJ method is just one theoretical framework within which Engineers can pursue the advancement of these capabilities, we argue that it provides a productive and nuanced way to discuss those capabilities. Outside the context of the E4SJ framework, for example, one might be able to motivate the development of certain technologies through their ability to enhance affinity, even if the target users whose affinity would be enhanced would be a group doing demonstrable societal harm, such as organized white supremacists. In contrast, operating within the context of the E4SJ framework encourages engineers to specify \textit{whose} affinity is being enhanced.

As such, in this paper, we use these capabilities, and the specific \textit{ways} in which the E4SJ framework suggests engineers seek to advance those capabilities, as a lens for analyzing the state of AI-HRI research. As we will show, the field of AI-HRI is at once has the \textit{potential} for strong alignment with work being done in AI-HRI, yet in \textit{practice} the motivations of AI-HRI research differ considerably. 

That is, on the one hand, \textit{all} of the capabilities laid out by Nussbaum stand to align with AI-HRI solutions, but in practice, most AI-HRI research is not explicitly motivated by these sorts of capabilities, and the research that is capability-motivated seeks to advance a narrow set of capabilities, such as preventing harm, promoting social engagement, providing education, and promoting good health. This suggests that the space of capability-focused solutions explored by AI-HRI researchers is perhaps overly focused on a few goals at the expense of others, perhaps due to the particular axiologies and lived experiences that are commonplace amongst AI-HRI researchers. 

Moreover, as we will show, the \textit{way} in which capabilities are typically advanced is misaligned with the community-focused approach that is proposed by the E4SJ framework.
Even when AI-HRI researchers produce technical advancements oriented around facets of, say, interaction, their solutions do not typically directly focus on the interaction needs of particular communities that are otherwise inequitably stymied by structural forces. And while we will not deeply discuss this point in this work, neither do researchers tend to include members of those communities in their research teams or even explicitly build off of work that does include and engage with members of those communities. 

Accordingly, we believe the E4SJ approach stands to address these shortcomings in the AI-HRI field's research practice. Specifically, we believe that the key human capabilities delineated by Nussbaum, when paired with a community-centered view of engineering as suggested by the E4SJ framework, can serve as guiding principles for the field, helping us to better gauge the promise of solutions being suggested by ourselves and others in our community from a social justice perspective.


To understand the extent to which the AI-HRI community is engaging in research practices aligned with principles of Engineering For Social Justice, we qualitatively analyzed the papers published at AI-HRI 2020. In the rest of the paper, we discuss the results of this analysis, and conclude with a vision for the future of AI-HRI.

\section{Method}
 For each paper, the paper authors assessed whether it (1) identified a specific community who it tried to help, (2) whether it expressed a motivation aligned with one of the ten key human capabilities, and (3) whether it expressed a motivation of performing some task \textit{in a way} that aligned with one of the ten key human capabilities, even if the main purpose of the robot did not. While these decisions were made subjectively, the authors did their best to try to make them systematically and generously, looking for any indication that could be used to justify inclusion in one of the categories on the basis of one of the criteria above.

\begin{figure*}[!ht]
    \centering
    \includegraphics[width=\linewidth]{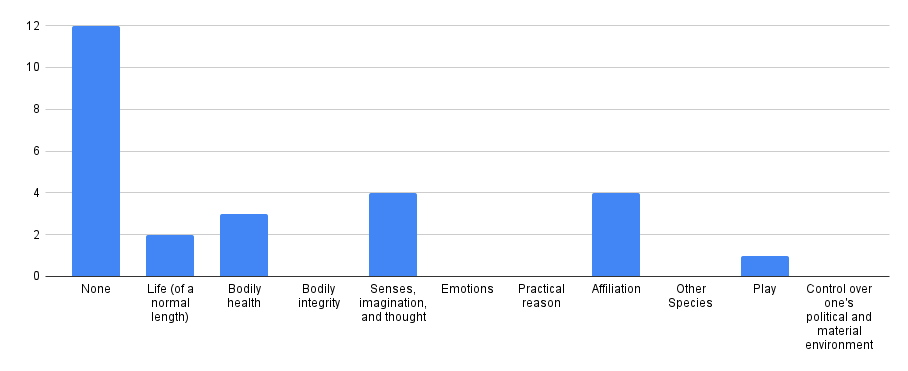}
    \caption{Papers engaging with each of Nussbaum's 10 human capabilities. Papers engaging with multiple capabilities were multiply counted. Papers indirectly engaging with a capability were counted for the purpose of this visualization as fully engaging.}
    \label{fig:chart}
\end{figure*}

\section{Results and Discussion}
Seventeen research papers were presented at AI-HRI 2020 (not counting four ``tool papers"). Of these, we identified four papers as clearly specifying a user population other than ``engineers", ``HRI researchers", or ``people who happen to be interacting with a robot in some un(der)specified and mysterious environment". The four papers that \textit{did} specify intended beneficiary communities presented technical advancements designed to help three user groups: (1) older adults~\cite{reneau2020supporting,wilson2020knowledge}, (2) children undergoing medical procedures~\cite{foster2020towards}, and (3) medical personnel supporting stroke victims~\cite{pourebadi2020stroke}. Based on even this limited information alone, it is interesting to see that when authors  thought about the communities their technology was intended to help, they directly or indirectly selected communities that were societally disadvantaged in some way; children, the elderly, and those helping these groups.

Of these seventeen papers, five engaged with at least one of Nussbaum's 10 human capabilities (Fig.~\ref{fig:chart}), and most of these five addressed multiple capabilities. 
Four of the five papers presented technologies that facilitated \textit{senses, imagination, and thought} in the sense of helping interactants avoid pain. One of these~\cite{foster2020towards} was also designed to promote avoidance of pain \textit{in a way} that also facilitated \textit{play}. Three of these four papers were among those that specified intended beneficiary communities~\cite{reneau2020supporting,foster2020towards,pourebadi2020stroke} (cf.~\citet{jeong2020face}). 
Four of the five papers presented technologies that either directly or indirectly facilitated \textit{affiliation}, in the sense of encouraging empathy and dignity; \citet{pourebadi2020stroke} and \citet{jeong2020face}'s approaches were designed with this capability in mind; \citet{reneau2020supporting} and \citet{wilson2020knowledge}'s approaches were designed to achieve other goals \textit{in a way} that advanced this capability. Three of these four papers were among those that specified intended beneficiary communities~\cite{reneau2020supporting,wilson2020knowledge,pourebadi2020stroke} (cf.~\citet{jeong2020face}).
Three of the five papers presented technologies that facilitated \textit{bodily health}~\cite{reneau2020supporting,wilson2020knowledge,pourebadi2020stroke}, two of which we also categorized as facilitating \textit{life}~\cite{wilson2020knowledge,pourebadi2020stroke}. All three were among those that specified intended beneficiary communities. 

Of the remaining twelve papers, eleven had concrete motivations not captured by the E4SJ framework: five were motivated by an abstract desire for explainability, three by an abstract desire for trustworthiness, and three by other abstract desires surrounding robot perception, cognition, and behavior modeling. While all of these types of approaches have the \textit{potential} to help create a more equitable society, the lack of articulation of an intended beneficiary community (and thus, subsequently, a lack of specification for how that community was intended to be helped by the technology) evoke a dangerous perspective in which these advances are seemingly cast as beneficial in and of themselves. 
While it is true that \textit{explainability} and \textit{trustworthiness} are admirable goal in some contexts (e.g., a robot that shares critical systemic knowledge with undocumented communities), these principles can be rendered dangerous when recontextualized into domains in which explanation-generation and trust-building mechanisms are deployed in order to coerce compliance with existing state power structures (e.g., robots deployed for the purpose of surveillance or oppression by corporate or state actors (e.g., police)). 
Similarly, while \textit{efficiency} can be an admirable goal in the contexts of making robots affordable for low-income communities, serving a greater number of hospitalized children, or enhancing disabled users' mobility, in many of the domains described in the analyzed papers, increased efficiency would primarily stand to benefit the wealthy executives and shareholders who may be exploiting the labor of those interacting with the robot. A social justice oriented approach to increasing efficiency in warehouse environments would need to be motivated by a community-provided efficiency concern grounded in one of Nussbaum's 10 human capabilities. For example, such an approach might be grounded in factory workers' fears that robots introduced into the workplace would decrease their efficiency in the sense that their specialized skills could go to waste~\cite{meissner2020friend}. This concern could be justified through its grounding in dignity, autonomy, and \textit{affiliation} that would need to be addressed in particular ways. 
Indeed, there is good reason to be skeptical of blind emphasis on metrics such as efficiency, effectiveness, and transparency, which are traditionally centered by neoliberal axiologies and theories of value~\cite{power2003evaluating,van2014neoliberal}.


\section{Conclusions: Envisioning a Social Justice oriented AI-HRI}

It is not our intent to imply that the papers we've chosen to (implicitly) critique are of poor quality or unethical. 
None of the papers published at AI-HRI last year were actively malicious or anti-social justice\footnote{Cp. recent papers~\cite{bordbar2021roman} published in the HRI community that have presented technologies actively intended to empower racist, violent institutions, without taking into account the perspectives of those likely to be targeted by the proposed technologies.}. Neither is it our intent to imply that they do not stand to address key capabilities from Nisbaum's taxonomy in some way. 
And moreover, most of the critiques leveled in this paper can be readily applied to the authors' own papers. 
Rather, we suggest that if we want to \textit{ensure} that our technologies are actually helping to build an equitable future, rather than simply helping those who are already socially and economically empowered, we should cultivate a culture of careful reflection in which we do our best to thoughtfully articulate answers to key engineering questions. Who is our technology actually intended to help? Whose capabilities (and which of their capabilities) are prioritized by our research efforts? And do our technologies actually help advance those capabilities?

Moreover, while only briefly touched on in this paper, we intend to suggest that our community should consider how the success of our attempts to advance key human capabilities \textit{equitably} are mediated by the other Engineering for Social Justice criteria: 
\begin{enumerate}
    \item What risks and harms are imposed by our technologies? How do our technologies increase opportunities and resources for our intended beneficiary communities? 
    \item How do our technologies politically empower communities? 
    \item What are the structural conditions that constrain the opportunities, desires, and aspirations or their intended beneficiary communities (both in terms of why technologies for \textit{those} communities are well-justified, and in terms of how our technologies stand to subvert those limitations)? 
    \item And finally, how are our proposed technologies grounded in contextual listening to communities' stories, values, and desires? 
\end{enumerate}
Asking ourselves these questions during our research process is a necessary step not only for ensuring that our engineering education efforts (for those of us teaching at universities) encourage students to engage in equitable and societally beneficial engineering practices, but also for ensuring that the technical advances we present at symposia like AI-HRI are truly advances as far as our society is concerned. A collective effort to re-focus on these types of questions may also lead us to reconsider the way that we read, interpret, enact, write our professional codes of ethics,  
We also admit that there are legitimate drawbacks to the approach proposed in this paper, and that the success of the E4SJ lens depends on the precise manner in which it is employed. A straightforward step that AI-HRI researchers could take to use the E4SJ criteria when motivating their work is to be specific about the specific community they are trying to help, and the specific capabilities their work is intended to advance for those communities. Even for highly theoretical work, researchers could give examples of communities their research would be expected to benefit and in what ways. However, this approach has some clear problems. 

In particular, the E4SJ criteria clearly center \textit{contextual listening} to communities. Mere speculation about the potential benefits of one's work to a particular community without talking to that community runs the risk of painting a human-centered veneer over one's research without doing the work of actually assessing the alignment between research and communities' self-expressed needs, values, and priorities (cf. recent critiques of ostensibly human-centered AI initiatives~\cite{alkhatib2019,le2020we}). Researchers doing foundational theoretical work cannot be expected to do deep participatory design work with specific communities, and there should be no expectation that their work should be immediately deployable in today's communities. And in fact, some have argued that doing participatory research on technologies that cannot be effectively and immediately deployed could be actively harmful, as at worst it could result in the deployment of technologies that are harmful due to that same nascent status, and at best, could result in wasting the time of communities without helping them. However, theoretical researchers could at least cite the work of others who \textit{have} done the work of documenting communities' needs, values, and priorities. 

Another concern is that researchers could use this type of justification to highlight potential pro-social uses of proposed technologies while ignoring potentially harmful dual-uses by other communities. This motivates a need for researchers to more broadly consider in their papers the wide range of uses potential technologies might have, both positive and negative (cf. recent discussions of such sections in NeurIPS papers~\cite{gibney2020battle,gibney2020ai,crawford2019ai}).

Finally, we conclude by envisioning in the section below what a possible future might look like, in which the AI-HRI community aggressively pursued a collective research program grounded in Social Justice, working to develop technical robotics advances that advanced key human capabilities for societally disadvantaged communities.

\subsection*{\textit{Speculative Exercise:} AI-HRI 2022 List of Accepted Papers (Titles Only)}

    \begin{enumerate}
    \item Facilitating \textbf{life}:  autonomous robot distribution of blankets to \textbf{homeless people} in public parks.
    \item Facilitating \textbf{bodily health}: socially assistive robots for encouraging exercise therapy participation in \textbf{older adults}.
    \item Facilitating \textbf{bodily integrity}: social robots for helping \textbf{sex workers} safely report abuse suffered at the hands of law enforcement.
    \item Facilitating \textbf{senses, imagination, and thought}:  \textbf{literacy tutoring robots} for \textbf{students from oppressed racial groups} attending underfunded segregated schools.
    \item Facilitating \textbf{emotions}:  conversational agents providing a safe sharing environment for \textbf{LGBT+ teenagers}.
    \item Facilitating \textbf{practical reason}: robot-led goal reflection with \textbf{first generation college students}.
    \item Facilitating \textbf{affiliation}: building cultivating environments for \textbf{women in STEM} with sexism-rebuking robots.
    \item Facilitating \textbf{connections to other species}: forest terrain adaptation algorithms for \textbf{robotic wheelchair users}.
    \item Facilitating \textbf{play}:  bilingual robots encourage structured play with \textbf{immigrant children}.
    \item Facilitating \textbf{control over one's environment}: building trustworthy robots to encourage census participation in \textbf{undocumented communities}.
    \end{enumerate}

\section*{Acknowledgements}
This work was funded in party by NSF grant IIS-1909847. We would like to thank Qin Zhu, Zhao Han, Terran Mott, and the paper reviewers for their insightful comments.

\bibliography{main}

\end{document}